\documentclass[10pt,twocolumn,letterpaper]{article}

\usepackage{ijcb}
\usepackage{times}
\usepackage{epsfig}
\usepackage{graphicx}
\usepackage{amsmath}
\usepackage{amssymb}

\usepackage{times}
\usepackage{epsfig}
\usepackage{graphicx,paralist,booktabs,tabu}
\usepackage{amsmath}
\usepackage{amssymb}
\usepackage{graphicx}
\usepackage{subcaption,comment}
\usepackage[sort&compress,numbers]{natbib}
\usepackage{bbm}
\usepackage[pagebackref=false,breaklinks=true,letterpaper=true,colorlinks,citecolor=green,bookmarks=false]{hyperref}
\newcommand\blfootnote[1]{%
  \begingroup
  \renewcommand\thefootnote{}\footnote{#1}%
  \addtocounter{footnote}{-1}%
  \endgroup
}

\ijcbfinalcopy 

\def\ijcbPaperID{50}  

\ifijcbfinal\fi
\begin{document}

\title{ Improved Presentation Attack Detection Using Image Decomposition
}

\author{Shlok Kumar Mishra  \textsuperscript{ \rm 1,\rm 2*}
Kuntal Sengupta  \textsuperscript{\rm 2},
Wen-Sheng Chu  \textsuperscript{\rm 2},\\
Max Horowitz-Gelb  \textsuperscript{\rm 2},
Sofien Bouaziz  \textsuperscript{\rm 2},
David Jacobs\textsuperscript{\rm 1} 
\\
\textsuperscript{\rm 1}University of Maryland, College Park,
    \textsuperscript{\rm 2}Google Research,\\
\texttt{\{shlokm,dwj\}@cs.umd.edu} 
}
\maketitle
\thispagestyle{empty}

\begin{abstract}
   Presentation attack detection (PAD)  is a critical component in secure face authentication.  
We present a PAD algorithm to distinguish face spoofs generated by a photograph of a subject from live images.  Our method uses an image decomposition network to extract albedo and normal.  The domain gap between the real and spoof face images leads to easily identifiable differences, especially between the recovered albedo maps.  We enhance this domain gap by retraining existing methods using supervised contrastive loss. 
We
present empirical and theoretical analysis that demonstrates that contrast and lighting effects can play a significant role in PAD; these show up particularly in the recovered albedo.   Finally, we demonstrate that by combining all of these methods we achieve state-of-the-art results on both intra-dataset testing for CelebA-Spoof, OULU, CASIA-SURF datasets and inter-dataset setting on SiW, CASIA-MFSD, Replay-Attack and MSU-MFSD datasets.     
\end{abstract}

\section{Introduction}
\blfootnote{* Work was done when Shlok was interning at Google Research.}

Recently, face recognition has been widely adopted for applications such as phone unlock and mobile payments.  
During a secure face authentication session, a presentation attack detection (PAD) system is tasked with distinguishing between a real human face and a face in a presentation attack instrument (PAI) in front of the camera. Examples of PAI includes printed paper attacks, replay attacks on displays, etc.
Commercial PAD methods rely on a depth sensor and active near-infrared illumination to increase the robustness of the system.  
This makes these methods unsuitable for low-cost mobile devices with a single RGB camera.   

\begin{figure*}[t]
  \centering
  \newlength\ww \setlength{\ww}{60pt}
  \begin{tabular}{c@{}c@{}c@{}c@{}c@{}c@{}c@{}c@{}}
    Image & Albedo & Normal & Depth & Image & Albedo & Normal & Depth \\
    \includegraphics[width=\ww]{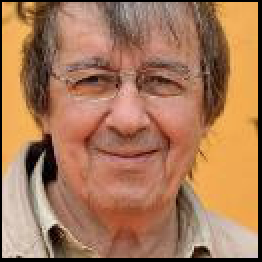} &
    \includegraphics[width=\ww]{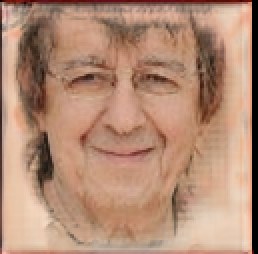} &
    \includegraphics[width=\ww]{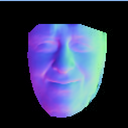} &
    \includegraphics[width=\ww]{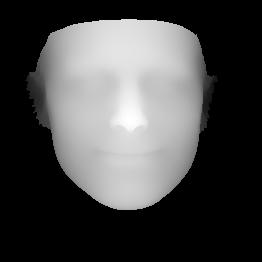} & 
    \includegraphics[width=\ww]{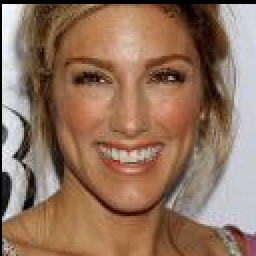} &
    \includegraphics[width=\ww]{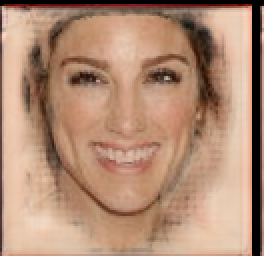} & 
    \includegraphics[width=\ww]{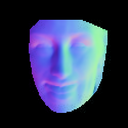} & 
    \includegraphics[width=\ww]{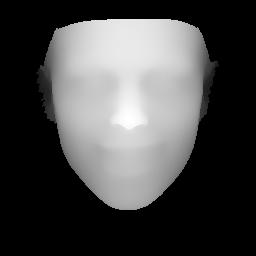} \\
    
    \includegraphics[width=\ww]{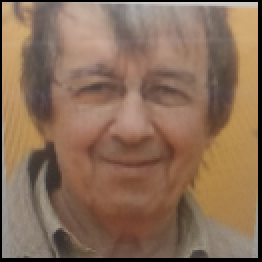} &
    \includegraphics[width=\ww]{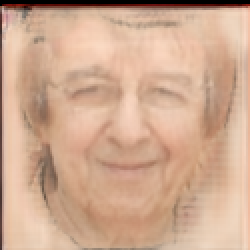} &
    \includegraphics[width=\ww]{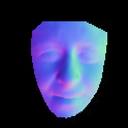} &
    \includegraphics[width=\ww]{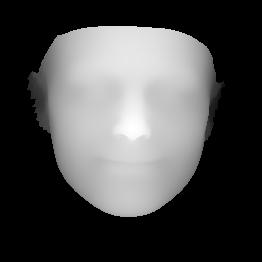}& 
    \includegraphics[width=\ww]{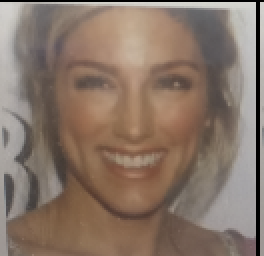} &
    \includegraphics[width=\ww]{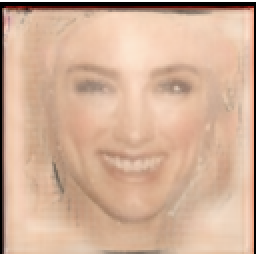} &
    \includegraphics[width=\ww]{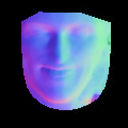} & 
    \includegraphics[width=\ww]{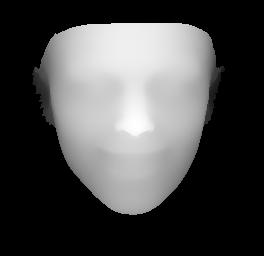}\\
  \end{tabular}
  \caption{
  Live images ({\em top row}) and spoof images ({\em bottom row}) show similar reconstructed depth maps and surface normal.
  Albedo, on the other hand, looks different for spoof images and live images. Albedo helps in capturing the contrast difference resulting from different lighting interactions that happen on the spoof images. 
  Here, we propose to use albedo as additional cue for detecting spoofs in addition to using depth maps.}
  \label{fig:teaser}
\end{figure*}

In recent years several handcrafted features   \cite{Boulkenafet2015FaceAB,boulkenafet2015face,Peixoto2011FaceLD,Patel2016SecureFU,Komulainen2013ContextBF,liu2015faceattributes} and deep learning-based approaches \cite{Yu2020AutoFasSL,George2019DeepPB,Gan20173DCN,Atoum2017FaceAU,Yang2019FaceAM,Jourabloo2018FaceDA,liu2015faceattributes,Qin2020LearningMM} have been designed for PAD. Recent methods  have  predicted  depth  signals from a single RGB image \cite{yu2020searching,wang2020deep,qin2019learning,yu2020multi} using PRNet \cite{feng2018prn}.
However, the reconstructed depth often looks similar for the real and the spoof images. Hence these methods hardcode the depth of spoof image to be zero. We on the other hand argue that face albedo is better suited for PAD. This is demonstrated in Figure \ref{fig:teaser} where the surface-normal (normal) and the depth maps are visually similar for real images and for spoof images.  The albedo map is visually different, motivating us to primarily condition our classification network on the face albedo. In contrast to depth we don't need to hardcode the albedo for real and spoof images as they are distinctively different. Also, estimating albedo doesn't require a expensive depth sensor hardware, hence making it suitable for low-cost mobile devices.

\begin{figure*}[t]
  \centering
    \includegraphics[scale=0.42]{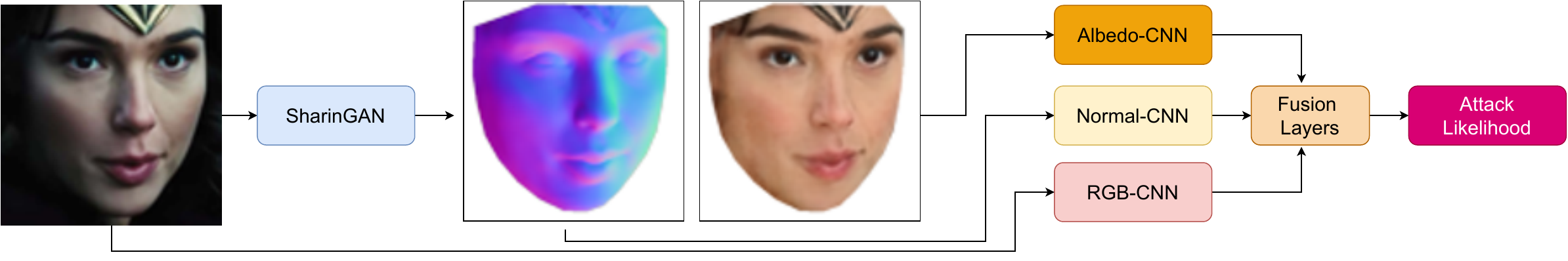}
    \caption{{\bf }IDCL network: We propose IDCL network to use image decomposition \cite{Pnvr2020SharinGANCS} for the task of PAD. We have three different networks corresponding to three different modalities i.e. RGB, albedo and normal. Empirically and visually, the albedo performs on-par in detecting spoofs compared to normal and depth maps.}
    \label{fig:IDCL}
\end{figure*}

Our emphasis on albedo is motivated by differences in contrast and lighting between real and spoof images, affecting the recovered albedo.  
We show empirically that real and spoof images have significantly different contrast.  In fact, removing this contrast difference results in a drastic drop in spoof detection performance by the state of the art approaches such as Central Difference Convolution(CDCN) \cite{yu2020searching}.  Theoretically, we show that lighting differences between real and spoof images also create effects that will show up in recovered albedo.  


These differences allow us to take
advantage of the domain gap between real images and spoof images   \cite{yu2020searching,jia2020singleside}. To verify this, we train the image decomposition network with only real images. This creates a domain gap when classifying spoof images. 
 We use supervised contrastive losses \cite{khosla2020supervised} to further increase this domain gap. Supervised contrastive loss is a generalization of triplet losses that can work on large batches of negative samples. 
 In PAD, we apply contrastive learning on real and spoof images. This forces the network to push the feature embeddings of real and spoof images far from each other, accentuating the difference between the real and the spoof domains. 

We propose the IDCL architecture (Fig \ref{fig:IDCL}) to assess the value of albedo and normal. Initially, we pretrain a network using supervised contrastive loss. We use this as the initialization point for the RGB backbone in our IDCL architecture.  
Next, we add surface normal and albedo signals as input to the IDCL network and train it to classify spoofs.
 From our experiments on CASIA-SURF \cite{zhang2020casiasurf} and OULU-NPU \cite{OULU_NPU_2017}, we observe that albedo is a better cue and has more impact on performance as compared to the surface normal.
The primary motive of IDCL architecture is to understand the effects of albedo and normal for spoof detection and not to improve upon state-of-the-art methods.

To achieve state-of-the-art results we add supervised contrastive loss initialization and albedo as a cue 
to various methods on seven different datasets.
We start by adding contrastive loss and an albedo component to AENet \cite{CelebA-Spoof}. AENet has state-of-the-art results on the CelebA-Spoof dataset, which is the largest public dataset available for spoof detection. Our methods outperform AENet \cite{CelebA-Spoof}, and we achieve state-of-the-art results on CelebA-Spoof. We also add supervised contrastive loss initialization and albedo components on top of CDCN \cite{yu2020searching} and on top of the winners of Chalearn Challenge\cite{Liu_2019_CVPR_Workshops} and achieve state-of-the-art results on OULU-NPU \cite{OULU_NPU_2017} and CAISA-SURF\cite{zhang2020casiasurf} as well. We also show state-of-the-art results on intra-dataset protocol on SiW dataset \cite{6199754} and on CASIA-MFSD \cite{6199754}, Replay-Attack \cite{6313548} and MSU-MFSD \cite{7031384} datasets.


Finally, we provide a theoretical analysis of the effect of lighting on spoof images.  Spoof images can be affected by the interplay between two lighting conditions.  Consider, for example, the case of a paper attack in which spoofs are captured using mobile phones.  There is one lighting condition when the original image is captured, and another one when the spoof picture is captured with the phone. In Sec \ref{sec:theoritical_analysis} we show that this combination of lighting can create effects on the face not generally produced by a single lighting condition. Specifically, we show that this interplay introduces new, higher frequency artifacts that can corrupt the albedo, and can be used to identify spoofs.
 
{\bf Our contributions} can be summarized as follows:
\begin{compactenum}
    \item We study 
    the impact of using albedo maps, normal and depth information for the task of PAD. Through qualitative and quantitative experiments, we verify that albedo is the stronger cue for detecting spoofs. 
    \item To the best of our knowledge, our approach is the first to show the use of supervised contrastive loss for PAD. Supervised contrastive loss helps generate better initialization points that transfer well for spoof detection.  
    \item We achieve state-of-the-art results on OULU-NPU, CelebA-Spoof, SiW dataset, CASIA-MFSD, Replay-Attack and MSU-MFSD datasets. We also demonstrate that by adding albedo as conditioning to various  
    state-of-art methods, their performance can be improved.
     \item Through theoretical analysis and experiments, we demonstrate that contrast and lighting effects  
     can play a significant role in spoof detection. 
    
\end{compactenum}

\section{Related Work}
{\bf Presentation Attack Detection (PAD):}
 Traditionally PAD has been solved for RGB images by using various local descriptors and features such as HOG features \cite{Komulainen2013ContextBF}, SIFT features \cite{Patel2016SecureFU}, or SURF \cite{boulkenafet2015face}. For video spoofing datasets, where input are videos, various methods have been proposed that capture dynamic features such as texture \cite{Komulainen2013ContextBF}, motion  \cite{Siddiqui2016FaceAW} and eye blinks \cite{Pan2007EyeblinkbasedAI}. Recently deep learning-based methods have been proposed to identify spoofs. Here, a convolutional neural network (CNN) backbone is trained to extract features that can be used to identify spoofs \cite{yu2020searching,yu2020multi,Qin2020LearningMM,Zhang2020FaceAV,Wang_2020_CVPR,Stehouwer_2020_CVPR}. A few methods also use depth as a cue to identify spoof images  \cite{yu2020searching,Yu2020MultiModalFA}. Our work shows that albedo is a stronger signal to identify spoofs compared to depth or other geometric information extracted from an RGB image; this is a marked departure from these standard methods. Concurrent work \cite{Zhu_2021_CVPR} uses a similar idea and shows results on Forgery Detection using  deviation  from the  common texture as an important  feature. We on the other hand focus on spoof detection and using albedo as a cue.  \\  
 {\bf Image Decomposition:}
Traditionally geometry estimation methods for faces were based on 3D Morphable models (3DMM)  \cite{Blanz1999AMM,Laine2017ProductionlevelFP,Shu2017NeuralFE,Tappen2005RecoveringII,Tewari2017MoFAMD,Tran2017RegressingRA,Thies2019Face2FaceRF}. Recently, a few studies have demonstrated the effectiveness of CNN's to solve the problem of geometry estimation  
\cite{Tewari2017MoFAMD,Shu2017NeuralFE,Tran2018Nonlinear3F,Genova2018UnsupervisedTF,Liu20193DFM,Sengupta2018SfSNetLS}. Depth estimated using PRNet \cite{feng2018prn} has also been used as a cue to detect spoofs \cite{yu2020searching,jia2020singleside,Qin2020LearningMM}. 
In our work, we propose to use image decomposition to estimate albedo. We use SharinGAN  \cite{Pnvr2020SharinGANCS}, which is based on SFSNet  \cite{Sengupta2018SfSNetLS} to extract albedo features and surface normals from images and use them as signals to detect spoofs.\\
 {\bf Contrastive Learning:}
Contrastive learning-based methods have been recently gaining popularity for self-supervised learning  \cite{He2020MomentumCF,Chen2020ImprovedBW,Chen2020ASF,Hnaff2019DataEfficientIR} and supervised learning  \cite{khosla2020supervised}. We use supervised contrastive learning to increase the domain gap between spoof images and real images, which gives us a better initialization point for our networks. 

\vspace{-0.1in}
\section{Our method}  

\label{sec:domain_gap}
 Recent methods have taken advantage of the domain gap between real and face images to identify spoofs \cite{Jourabloo_2018_ECCV,jia2020singleside}. Potential reasons for this gap include color distortion, display artifacts, presenting artifacts and imaging artifacts as discussed by Face De-Spoofing  \cite{Jourabloo_2018_ECCV}.  We show that domain gap can also be caused by the contrast difference between the real and the spoof images and also by  
the presence of a linear gradient on the images which may cause these networks to learn sub-optimal representations. We discuss these issues in more detail in Section \ref{sec:analysis}.
Now we discuss our proposed approach, which amplifies the domain gap.

\begin{figure}[t]
    \centering
    \includegraphics[width=\linewidth]{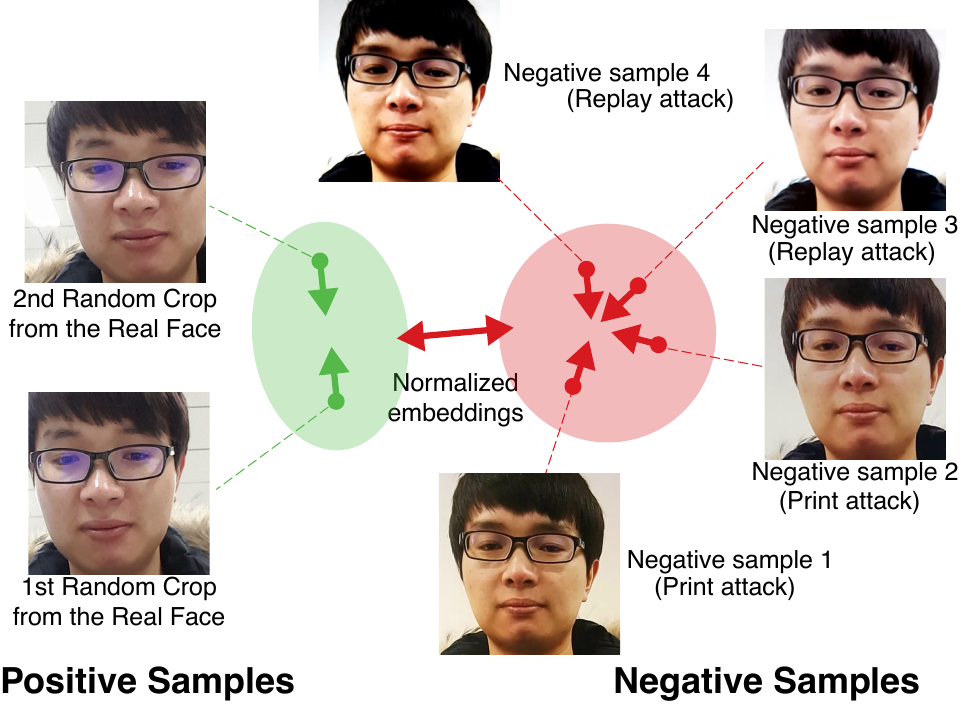}
    \caption{ We use supervised contrastive learning to increase the domain gap between spoof images and live images. In supervised contrastive learning the negative sample comes from different class from the positive class.
    If the positive sample is from the real image, the negative sample would be from a spoof image and vice-versa. Hence a contrastive learning approach will push real images and spoof images away from each other, thus increasing the domain gap.
    }
\end{figure}

\subsection{Amplifying the domain gap with image decomposition:} 

 To calculate geometric features from real images, we use SharinGAN \cite{Pnvr2020SharinGANCS}, which computes normal, albedo and lighting. SharinGAN \cite{Pnvr2020SharinGANCS} is based on SFSNet \cite{Sengupta2018SfSNetLS} with additional components that reduce the domain gap between synthetic and real training data.
Next, we use the normal and albedo computed by SharinGAN as input to our network, along with the original RGB.  We train three networks independently, one for each modality, i.e RGB, albedo and normal and fuse them in later layers, as shown in Fig.~\ref{fig:IDCL}.
 
\textbf{Amplifying the domain gap in image decomposition:} 
 Domain gaps between training and testing data have been a long-standing issue in Computer Vision \cite{Volpi2018GeneralizingTU,Mahajan2020DomainGU,Mancini2020TowardsRU,albuquerque2020generalizing}. However, this gap can be advantageous in our case, if it can be used to cause reconstruction methods to perform differently on real and spoof images. 
   To achieve this, we train SharinGAN \cite{Pnvr2020SharinGANCS} using only real images from the Spoofing dataset. By doing this, the network will not have seen the spoof images during training, and hence the network will produce more accurate image decomposition  
   for real images as compared to spoof images. 
We call this network SharinGAN-Real and its impact is shown in Table \ref{tab:OULU}. 

\subsection{Increasing the domain gap with contrastive loss}  
\label{sec:contrastive}
Traditionally, methods such as adversarial loss have been used to create latent features that have the same distribution for both domains.  We instead use the contrastive loss to widen the gap between these distributions.  

Contrastive methods are generalizations of triplet losses \cite{khosla2020supervised} and have been shown to work very well in Supervised and Self-Supervised Learning (SSL) methods \cite{He2020MomentumCF,Chen2020ImprovedBW}. We will first discuss the details of supervised contrastive loss and then its application to the spoof detection problem. 

\textbf{Supervised contrastive loss}: 
Supervised contrastive loss \cite{khosla2020supervised} imposes a constraint that normalized embeddings from the same class are closer than the ones from different classes.
Contrastive losses consist of two “opposing forces”: for a given anchor point, the first force pulls the anchor closer to positive examples in the representation space, and the second force pushes the anchor farther away from negative points. 
In the supervised contrastive loss, negative samples come from different classes than the positive class. 

 
 In our case, the positive samples are augmentations of the same face and negative samples would be any spoof image. We perform augmentation by taking two random crops from the same face.
 Hence the contrastive training method will force the network to push the representations of real images and spoof images far from each other, thus increasing the domain gap between real and spoof images. 
 


\textbf{IDCL: Image Decomposition and Contrastive Loss}
Now we discuss how we combine the image decomposition method SharinGAN \cite{Pnvr2020SharinGANCS} and supervised contrastive loss. 
Our proposed architecture IDCL (image decomposition and contrastive loss)  as shown in Fig \ref{fig:IDCL} has three branches for the three modalities, RGB, albedo and normal. First, we pretrain a randomly initialized neural network by supervised contrastive loss on the RGB images. We use this pretrained network and initialize the RGB branch of the IDCL network from this pretrained point. Next, we use the output of the SharinGAN (i.e albedo and normal components), and feed these components and the RGB frames as inputs to three different branches in our IDCL network. We also aggregate information from all the preceding layers in their final layers as was done in \cite{Liu_2019_CVPR_Workshops}. Note that we don't set the depth map of spoof image to be zero which is a common pratice in state-of-the-art methods; instead we use the output of PRNet \cite{feng2018prn} directly for both real and spoof images.  We then train our IDCL network in an end to end fashion using all these three modalities. More details of the the IDCL architecture can be found in supplementary.


\section{Results}
We start by briefly describing the datasets used in our experiments.  Then we show that using supervised contrastive losses can help in achieving better initialization and can boost performance by a fair margin. We then present results supporting our hypothesis that albedo is a strong cue for detecting spoofs. We demonstrate this across  two datasets, CASIA-SURF \cite{zhang2020casiasurf} and OULU-NPU \cite{OULU_NPU_2017}. Then we show the impact of the domain gap by only using real images while training image decomposition methods. Finally we verify that adding albedo as a signal to OULU-NPU,CASIA-SURF and CelebA-spoof helps us in achieving state of the art results on these datasets.
\subsection{Datasets}
\textbf{CelebA-Spoof} \cite{CelebA-Spoof}: CelebA-Spoof is a recently released dataset and the biggest open source spoof corpus. CelebA-Spoof \cite{CelebA-Spoof} has 625,537 samples of 10,117 subjects, and is significantly bigger than other Anti-Spoofing datasets. In total there are 8 scenes (2 environments * 4 illumination conditions) and more than 10 sensors. It also has 10 spoof type annotations and 40 attribute annotations closely following the original CelebA-dataset \cite{liu2015faceattributes}. 

\textbf{OULU-NPU:} OULU-NPU is a high quality dataset \cite{OULU_NPU_2017} that contains four protocols to test the generalization capability of anti-spoofing systems. 

\textbf{CASIA-SURF:} CASIA-SURF \cite{zhang2020casiasurf} is a large scale dataset with 21,000 videos. Each sample of CASIA-SURF has three modaliltes ( i.e RGB, Depth and NIR).

\textbf{SiW:} SiW dataset is designed for cross-type testing for unseen attacks.  We show cross-domain testing results on SiW dataset.

\textbf{CASIA-MFSD \cite{6199754}, Replay-Attack \cite{6313548} and MSU-MFSD \cite{7031384}:} 
CASIA-MFSD \cite{6199754}, Replay-Attack \cite{6313548} and MSU-MFSD \cite{7031384} datasets contain low-resolution videos which are used for cross domain testing.

\begin{table*}
    \caption{We show results on CASIA-SURF. The first row shows the results of using ground truth depth and NIR information; we can see that this solves the problem quite well. The second row shows the baseline using only RGB frames. Using supervised contrastive loss helps us in achieving 10.21\% improvements over the RGB baseline. We can see that after using SharinGAN, we improve by 59.09\% over our RGB baseline.  }
    \label{tab:CASIA}
    \centering
    \begin{tabu} to \linewidth {lccc} 
        \toprule
        {\bf Method} & TPR@FPR=10-2 & TPR@FPR=10-3 & TPR@FPR=10-4 \\ 
        \midrule
        Using RGB/NIR/depth & 0.9997 &0.9967 &0.6850 \\ 
        
        Using Only RGB & 0.4646 & 0.1299 & 0.0651 \\
        Supervised Contrastive (Ours) & 0.5119 & 0.1676 & 0.1151 \\
        IDCL (Ours) & 0.7391 & 0.3373 & 0.1126 \\
        
        \bottomrule
    \end{tabu}
    
\end{table*}

\begin{table*}
    \caption{We present the results on OULU-NPU for Protocol 1. We observe that after using SharinGAN we improve by 44\% on our RGB baseline. Using supervised contrastive we improve by 15.58\%  over the RGB baseline.}
    \vspace{-1.5ex}
    \label{tab:OULU}
    \centering
    \begin{tabu} to \linewidth {lccc}
        \toprule
        {\bf Method} & TPR@FPR=10-2 & TPR@FPR=10-3 &TPR@FPR=10-4 \\
        \midrule
        RGB baseline & 0.43 & 0.05 & 0.05 \\
        RGB with Supervised Contrastive loss & 0.53 & 0.25 & 0.25 \\
        RGB and SharinGAN (using albedo and normal) & 0.53 & 0.24 & 0.24 \\
        RGB and SharinGAN-Real (using albedo and normal) & 0.65 & 0.39 & 0.39 \\
        \bottomrule
        
    \end{tabu}
\end{table*}
\begin{table}[b]
    \caption{Results on CASIA-SURF using albedo, depth and IR information. The metric used is TPR@FPR=threshold.}
    \label{tab:CASIA_sota}
    \centering
    \begin{tabu} to \linewidth {lccc} 
        \toprule
        {\bf Method} & thr=1e-2 & thr=1e-3 & thr=1e-4 \\ 
        \midrule
        Baseline\cite{Liu_2019_CVPR_Workshops} & 100.0 & 100.0 &.9987 \\ 
        Ours & 100.0 & 100.0 &.9998 \\ 
        
        \bottomrule
    \end{tabu}
\vspace*{-0.17in}
\end{table}
\begin{table*}
    \caption{Ablation results dropping albedo and normal for both datasets. We first removed the normal maps, and we see a performance drop of 7.9\% for CASIA-SURF and 9.09\% for OULU. Next, we remove the albedo components and we see a more significant drop of 25.8\% for CASIA-SURF and 18.8\% for OULU. This shows that albedo is a bigger cue, and dropping albedo impacts the performance more than dropping normal.}
    \label{tab:Ablation}
    \centering
    \begin{tabu} to \linewidth {lcccc}
        \toprule
        {\bf Method} & Dataset & TPR@FPR=10-2 & TPR@FPR=10-3 &TPR@FPR=10-4 \\
        \midrule
        IDCL & OULU &  0.65 & 0.39 & 0.39 \\
        Albedo removed & OULU & 0.55 & 0.31 & 0.31 \\
        Normal removed & OULU & 0.60 & 0.53 & 0.53 \\
        \midrule
        IDCL & CASIA-SURF & 0.73 & 0.33 & 0.11 \\
        Albedo removed & CASIA-SURF  &  0.57  & 0.33  & 0.10  \\
        Normal Removed & CASIA-SURF & 0.68 & 0.27 & 0.10  \\
        \bottomrule
    \end{tabu}
\end{table*}
\begin{table*}
    \caption{Results on CelebA-Spoof after adding Albedo to AENet \cite{CelebA-Spoof}. We can see that adding albedo component improves performance over baseline AENet \cite{CelebA-Spoof}. Supervised contrastive loss initialization helps in improving the results even further as shown in last row.}
    \centering
    \begin{tabu} to \linewidth{l *{3}{X[c 1.3]}*{3}{X[c 0.8]}}
        \toprule
        {\bf Method} & TPR@FPR=1\% & TPR@FPR=0.5\% &TPR@FPR=0.1\%&APCER & BPCER & ACER  \\
        \midrule
        Baseline AENet  \cite{CelebA-Spoof} & 98.9 & 97.8 & 90.9 & 4.62 & 1.09 & 2.85 \\
        Albedo-SharinGAN (Ours) & \textbf{99.1} & \textbf{98.1} & \textbf{91.2} & \textbf{4.31} & \textbf{0.93} & \textbf{2.62} \\
        + Contrastive Loss (Ours) & \textbf{99.3} & \textbf{98.4} & \textbf{91.5} & \textbf{3.79} & \textbf{0.81} & \textbf{2.30} \\
        \bottomrule
    \end{tabu}
    \label{tab:AENet_sota}
\end{table*}

\begin{table*}
    \vspace{-0.5ex}
     \caption{We show the impact of removing contrast as a cue from spoof detection systems. We can see that after removing contrast, we see a performance drop in spoof detection on CDCN\cite{yu2020searching}. Hence to capture this contrast cue and other lighting cues, we propose to use albedo to detect spoof images. }
    \centering
    \begin{tabu} to \linewidth{l *{3}{X[c 1.3]}*{2}{X[c 0.8]}}
        \toprule
        {\bf Method} & Protocol & APCER & BPCER & ACER  \\
        \midrule
        CDCN  \cite{yu2020searching} & 1 & 0.4 & 0.0 & 0.2 
        \\
        CDCN contrast removed  \cite{yu2020searching} & 1 & 4 & 3 & 3.5 \\
        
       
        
        \bottomrule
    \end{tabu}
   
    \label{Tab:histogram}
    
\end{table*}

\begin{table}
    \centering
    \caption{Results on OULU-NPU \cite{OULU_NPU_2017} on  protocols P=1,2,3,4. We add an albedo component to the CDCN \cite{yu2020searching} architecture and achieve state-of-the-art results.
    Our experiment is denoted as ``SharinGAN+CDCN'' as the last row, where we add albedo component from SharinGAN and add it to CDCN \cite{yu2020searching}.
    }
     \label{tab:SOTA_OULU}
    \small
    \begin{tabu} to \linewidth {@{}X[l 0.01] X[l 3.2] X[c 1.3] X[c 1.3] X[c 1.3]@{}} 
        \toprule
        P & {\bf Method} & APCER & BPCER & ACER \\
        \midrule
        1 & FaceDs \cite{Jourabloo2018FaceDA} & 1.2 & 1.7 & 1.5 \\
        1 & FAS-TD \cite{Wang2018ExploitingTA} & 2.5 & 0.0 & 1.3 \\
        1 & CDCN++ \cite{yu2020searching} (Baseline)(Reproduced) & 1.5 & 1.2 & 1.3\\
        
        1 & \textbf{NAS-FAS\cite{Yu2021NASFASSC}} & \textbf{0.4} & \textbf{0.0} & \textbf{0.2} \\
        
        1 & \textbf{SharinGAN+CDCN} & {0.8} & {0.7} & {0.7} \\
        1 & \textbf{+ Contrastive Loss} & {0.7} & {0.6} & {0.6} \\ 
        
        \midrule
        
        2 & GRADIANT \cite{boulkenafet2015face} & 3.1 & 1.9 & 2.5 \\
        2 & STASN \cite{Yang2019FaceAM} & 4.2 & 0.3 & 2.2 \\
        
        2 &  CDCN++ \cite{yu2020searching} (Baseline)(Reproduced) & 2.0 & 1.6 & 1.8 \\
        2 & {NAS-FAS\cite{Yu2021NASFASSC}}  & {1.5} & \textbf{0.8} & \textbf{1.2} \\
        2 & \textbf{SharinGAN+CDCN} & \textbf{1.5} & {1.4} & {1.4} \\
        2 & \textbf{+ Contrastive Loss} & \textbf{1.3} & {1.1} & \textbf{1.2} \\
        
        \midrule
        
        3 & FaceDs \cite{Jourabloo2018FaceDA} & 4.0$\pm$1.8 & 3.8$\pm$1.2 & 3.6$\pm$1.6 \\
        
        3 & STASN \cite{Yang2019FaceAM} & 4.7$\pm$3.9 & 0.9$\pm$1.2 & 2.8$\pm$1.5 \\
        3 & CDCN++ \cite{yu2020searching} (Baseline)(Reproduced) & 2.3$\pm$1.5 & 2.5$\pm$1.2 & 2.3$\pm$0.7 \\
        3 & {NAS-FAS\cite{Yu2021NASFASSC}}  & {2.1$\pm$1.3} & \textbf{1.4$\pm$1.1} & {1.7$\pm$0.6} \\
        3 & \textbf{SharinGAN+CDCN} & {2$\pm$1.5} & {2.1$\pm$1.1} &{ 2.0$\pm$0.7} \\
        3 & \textbf{+ Contrastive Loss} & \textbf{1.7$\pm$1.4} & {1.8$\pm$1.1} &\textbf{ 1.7$\pm$0.7} \\ 
         
        \midrule
        
       4&FAS-TD~ \cite{Wang2018ExploitingTA}&14.2$\pm$8.7 &4.2$\pm$3.8 & 9.2$\pm$3.4 \\
       4&STASN~ \cite{Yang2019FaceAM}&6.7$\pm$10.6 &8.3$\pm$8.4  &7.5$\pm$4.7 \\
       
       4& CDCN++ \cite{yu2020searching} (Baseline)(Reproduced)
       &4.5$\pm$3.4 &6.3$\pm$4.9  & 5.3$\pm$2.9 \\
       3 & {NAS-FAS\cite{Yu2021NASFASSC}}  & {4.2$\pm$5.3} & \textbf{1.7$\pm$2.6} & \textbf{2.9$\pm$2.8} \\
       4 & \textbf{SharinGAN+CDCN} & \textbf{4.0$\pm$1.6} & {6.0$\pm$4.8} & {5.0$\pm$2.8} \\ 
       4 & \textbf{+ Contrastive Loss} & \textbf{3.4$\pm$1.5} & {5.5$\pm$4.4} &{4.5$\pm$2.7} \\ 
     \bottomrule  
     \end{tabu}
\end{table}

\begin{table}
\centering
\caption{Results of intra testing on three protocols of SiW. } 
\resizebox{0.45\textwidth}{!}{
\begin{tabular}{|c|c|c|c|c|}
\hline
Prot. & Method & APCER(\%) & BPCER(\%) & ACER(\%) \\
\hline
        1&{CDCN++ }&0.07 &0.17 & {0.12} \\
        &\textbf{CDCN + SharinGAN(Ours)}& \textbf{0.04} & \textbf{0.12} & \textbf{0.6} \\
\hline
      2 &{CDCN++ } &0.00$\pm$0.00 &0.09$\pm$0.10  & {0.04$\pm$0.05} \\
      &\textbf{CDCN + SharinGAN(Ours)} & \textbf{0.00$\pm$0.00} & \textbf{0.09$\pm$0.10}  & \textbf{0.04$\pm$0.05} \\
\hline
       3&{CDCN++ } &1.97$\pm$0.33 &1.77$\pm$0.10  & 1.90$\pm$0.15 \\
       &\textbf{CDCN + SharinGAN(Ours)} &\textbf{1.43$\pm$0.30} & \textbf{1.54$\pm$0.10}  & \textbf{1.58$\pm$0.15} \\
\hline
\end{tabular}
}
\label{tab:SiW}
\vspace{-1.0em}
\end{table}

\newcommand{\tabincell}[2]{\begin{tabular}{@{}#1@{}}#2\end{tabular}}
\begin{table*}
\centering
\caption{AUC (\%) of the model cross-type testing on CASIA-MFSD \cite{6199754}, Replay-Attack \cite{6313548}, and MSU-MFSD \cite{7031384}}

\scalebox{0.7}{\begin{tabular}{|c|c|c|c|c|c|c|c|c|c|c|}
\hline
{Method} &\multicolumn{3}{c|}{CASIA-MFSD} &\multicolumn{3}{c|}{Replay-Attack }&\multicolumn{3}{c|}{MSU-MFSD} &{Overall} \\
\cline{2-10} &\tabincell{c}{Video} &\tabincell{c}{Cut Photo} &\tabincell{c}{Wrapped Photo} &\tabincell{c}{Video}&\tabincell{c}{Digital Photo}&\tabincell{c}{Printed Photo}&\tabincell{c}{Printed Photo}&\tabincell{c}{HR Video}&\tabincell{c}{Mobile Video} & \\
\hline
DTN ~\cite{liu2019deep}
& 90.0 & 97.3 & 97.5 & 99.9 & 99.9 & 99.6 & \textbf{81.6} & 99.9 & 97.5 & 95.9$\pm$6.2 \\
\hline
{CDCN }
& 98.48 & \textbf{99.90} & \textbf{99.80} & \textbf{100.00} & 99.43 & 99.92 & 70.82 & \textbf{100.00} & \textbf{99.99} & 96.48$\pm$9.64 \\
\hline
{CDCN++ }
& 98.07 & \textbf{99.90} & 99.60 & 99.98 & {99.89} & \textbf{99.98} & 72.29 & \textbf{100.00} & 99.98 & {96.63$\pm$9.15} \\
\hline
\textbf{SharinGAN + CDCN (Ours)}
& \textbf{98.88} & \textbf{99.90} & \textbf{99.80} & \textbf{100.00} & \textbf{99.91} & \textbf{99.99} & {73.89} & \textbf{100.00} & \textbf{99.99} & \textbf{96.71$\pm$9.27} \\
\hline
\end{tabular}
}
\label{tab:cross-type}
\end{table*}


\subsection{Results using the IDCL architecture}

In this section, we show the results of using image decomposition methods and contrastive learning. 

{\bf Using contrastive learning:}
The use of supervised contrastive loss helps in increasing the already existing domain gap between spoof and non-spoof images. To evaluate this we pretrain our network using supervised contrastive learning using only RGB images.  Then we use this pre-trained backbone and finetune for the task of spoof classification. We compare our results with an RGB baseline in which we train a randomly initialized network for the task of spoof classification as shown in Table  \ref{tab:OULU} and Table  \ref{tab:CASIA}. 
We can see from Table  \ref{tab:OULU} and Table  \ref{tab:CASIA} that adding contrastive loss helps in learning better representations. We can see that there is an improvement of around 15.58\% on the OULU dataset and 10.21\% for the CAISA-SURF dataset as compared to the RGB baseline. Hence, when only using RGB images we can see that pre-training the network with supervised contrastive loss helps in learning better representations. 

{\bf Using all three modalities:} 
We use a pretrained SharinGAN \cite{Pnvr2020SharinGANCS} on CelebA to extract depth and albedo information from images. 
Next, to accentuate the domain gap we train SharinGAN \cite{Pnvr2020SharinGANCS} using only real images from the spoofing datasets. This encourages SharinGAN to make better predictions for real images than for spoof images.
 We can see the impact of training the network using only real images in Table \ref{tab:OULU}. We can see that using SharinGAN-Real we are able to achieve performance improvements of 50.01\% over the RGB only baseline.  Hence this shows that training SharinGAN with real images and taking advantage of domain gap helps in learning better representations. We only verify this result for OULU-NPU and not for CASIA-SURF, because the quality of images is too low in CASIA-SURF to retrain image decomposition methods. In the case of CelebA-Spoof, there is no domain gap since SFSNet and SharinGAN have been trained on the CelebA dataset as the real-world dataset. We only illustrate results using normal, since depth maps and normals are principally very similar. 

\subsection{Ablation study on albedo and normal}
{\bf IDCL:}
One of the main hypotheses in our work is that albedo is a better cue for detecting spoofs as compared to depth and surface normals as shown in Fig  \ref{fig:teaser}. 
To examine this difference empirically, we perform ablation studies on normals and albedo. To do this, we drop one branch at a time from the IDCL architecture shown in  Fig  \ref{fig:IDCL}. We can see from Table  \ref{tab:Ablation} for both datasets dropping albedo results in significant drops in performance(25.8\% for CASIA-SURF and 18.8\% for OULU) and dropping normal results in a smaller drop in the performance(7.9\% for CASIA-SURF and 9.09\% for OULU). This suggests that albedo is a stronger cue for detecting spoofs in both the datasets. 

\vspace{-0.1in}
\subsection{State-of-the-art results using albedo and contrastive loss}
In addition to showing that albedo is a useful cue for detecting spoofs, we also show that adding albedo to existing methods can be used to get state-of-the-art results on high-quality datasets like OULU-NPU \cite{OULU_NPU_2017} and large datasets like CelebA-spoof \cite{CelebA-Spoof} and CASIA-SURF \cite{zhang2020casiasurf}. We also show that adding contrastive loss initialization can help in achieving even better results.
To show state-of-the-art results we add an albedo component to the baselines architectures. We follow CDCN \cite{yu2020searching} for OULU, AENet \cite{CelebA-Spoof} for CelebA-spoof \cite{CelebA-Spoof} and build on top of the winners of Chalearn-Challenge \cite{Liu_2019_CVPR_Workshops} for CASIA-SURF. 
We see that we can improve upon the performance of CDCN \cite{yu2020searching} across all four OULU protocols, as shown in Table \ref{tab:SOTA_OULU}. We are also able to improve upon the performance of CelebA \cite{CelebA-Spoof} as shown in Table \ref{tab:AENet_sota} and on CASIA-SURF as shown in Table \ref{tab:CASIA_sota}. In the case of CASIA-SURF these results are pretty saturated, and there's little room for improvement.
We also show cross-testing results on  MSU-MFSD \cite{Wen2015FaceSD}, Replay-Attack \cite{Chingovska2012OnTE} and CASIA-MFSSD \cite{Zhang2012AFA} as shown in Table \ref{tab:cross-type}. 
Similarly cross-testing on SiW dataset are shown in Table \ref{tab:SiW}.

 
 

\section{Analysis}
 \label{sec:analysis}
 In this section, we discuss contrast cues and theoretical analysis, considering the effect of ambient lighting on spoof images.
 We show some additional visual analysis which is produced by image decomposition methods in the supplementary material.

 \subsection{Occlusion based visualization}
\begin{figure}[b]
  \centering
  \newlength\wwz \setlength{\wwz}{60pt}
  \begin{tabular}{c@{}c@{}c@{}c@{}c@{}}
    Live Image & Occlusion & Spoof Image & Occlusion \\
    \includegraphics[width=\wwz]{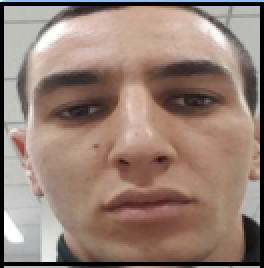} &
    \includegraphics[width=\wwz]{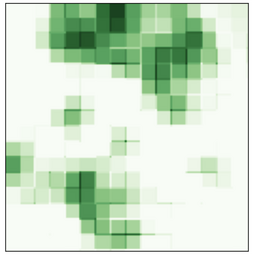} &

    \includegraphics[width=\wwz]{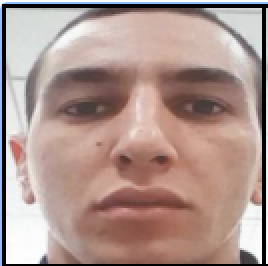} &
    \includegraphics[width=\wwz]{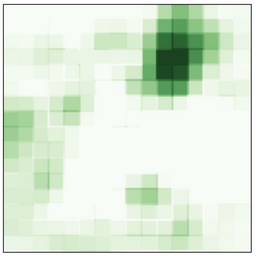} &
  \end{tabular}
  \vspace{-2.5ex}
  \caption{Visualization using occlusion based attribution.}
  \vspace*{-0.2in}
  \label{fig:albedo_occlusion}
\end{figure}

 To better understand our network features, we apply occlusion at different portions of the input image with a grey patch and monitor the classifier output.
Green patches in Fig \ref{fig:albedo_occlusion} indicates the regions where occlusion leads to a strong drop in activation in the feature map. 
More activation drop can be observed in the live image than in the spoof image, showing that albedo feature contains more texture information of the live image than of a spoof one.
 
\subsection{Contrast cues}
We show that contrast is a significant cue for spoofs in current state-of-the-art datasets  \cite{OULU_NPU_2017,zhang2020casiasurf}.  To verify contrast as a cue, we first calculated the mean and standard deviation of contrast on the OULU training set for real images and spoof images. We center the images to have zero mean and calculate the root mean square of pixel intensity, i.e. contrast value. For real images, the contrast was $81.64\pm14.97$, and for spoof images, the contrast was $72.36\pm13.91.$ We observe that there is a significant gap between the contrast of real images and spoof images. To remove contrast as a cue, we perform histogram equalization on the images. Next, we retrained the current state of the art method CDCN \cite{yu2020searching}, and we observe a significant drop in performances in Table  \ref{Tab:histogram}. 
This performance drop suggests that contrast as a cue is being used in current systems. Hence we propose to use albedo, which is especially well suited to help us in capturing this contrast difference. Complete table with results on all the protocols is shown in supplementary materials.

\subsection{Theoretical analysis of ambient lighting}
\label{sec:theoritical_analysis}

In this section, we consider the effect of ambient lighting on spoof images.  In a photograph of a face, appearance is influenced by the lighting.  In PAD, one captures a photograph of an existing photograph. Hence, the spoof image is also affected by a second lighting condition, as the Presentation Attack Instrument (the photograph in this case) itself reflects light.  We analyze a simple case to show how this second lighting condition can produce effects that could not have arisen in an original photograph of a face.  

It is well-known that an image of a convex Lambertian object is primarily determined by the low-frequency components of light, hardly reflecting any of the higher frequency components  \cite{basri2003lambertian,ramamoorthi2001efficient}.  Several previous works 
have applied this model to faces with excellent results, indicating that this is a good model in spite of the non-convex, non-Lambertian aspects of human faces  \cite{basri2003lambertian,zhang2003face,lee2005acquiring,wagner2011toward}.  When a photograph of a face is imaged, the light impinging on the photograph scales its intensities.  Below we show that in general, this can introduce higher frequency effects that were not present in the original photo.
\vspace{-0.15in}
\subsubsection{Analysis of a simple case}
\vspace{-0.07in}
In this section, we analyze the simplified case of a photo of a white Lambertian unit sphere.  Then we discuss the relevance of this result to our approach.

We first consider distant, environment map lighting.
Distant light can be represented as intensity as a function of direction.  We denote this function as $r(\hat{u})$, where $\hat{u}$ is a unit vector indicating direction.  With Lambertian reflectance, the irradiance, that is, the intensity of light reflected by a point on the sphere is:
\[
i(\hat{v}) = \int_{\hat{u}} r(\hat{u})\max(0, \hat{u}\cdot\hat{v}) d\hat{u}
\]
This is essentially convolution with a cosine filter that is clipped to be non-negative.  Analogous to the Fourier basis for signals in ${\cal R}^n$, the lighting and irradiance can be written as a linear combination of spherical harmonics.  The Funk-Hecke theorem is analogous to the convolution theorem, and tells us that for:
\[
r = \sum_{l=0}^\infty \sum_{m = -l}^l r_l^m Y_l^m
\]
we have
\[
i = \sum_{l=0}^\infty \sum_{m = -l}^l i_l^m k_l^0 Y_l^m
\]
Here $Y_l^m$ denote the spherical harmonics. The index $l$ indicates the order of the harmonic, corresponding to the frequency in the Fourier basis. Also, $r_l^m$ are coefficients of the spherical harmonic transform of $r$, and $k_l^0$ are coefficients of the transform of the convolution kernel, which is a clipped cosine function.  Since this is zonal, only coefficients with $m=0$ are non-zero.

For distant lighting,  \cite{basri2003lambertian} shows that $k_l^0 = 0$ for odd values of $l > 1$.  For even values, $k$ decays rapidly, and $k_l^0$ becomes quite small for $l>2$.  This implies that the image is largely determined by the first nine spherical harmonics of the lighting, of order 0, 1, and 2. When the lighting is not distant,  \cite{frolova2004accuracy} show that third order harmonics can also appear in $i$, but their effect becomes minimal when the light source is not very close to the object.  For example, when the distance to the light is eight times the sphere's radius, these effects become almost imperceptible.

Next, we consider the effect of light impinging on a printed photograph that is being imaged to create a spoof.  We assume that the photo is matte (eg., Lambertian) and planar.  If this is not the case, geometric distortion effects will provide further cues that can be used to detect a spoof.
If the photograph is flat, and the light sources are distant, then the light's intensity is completely uniform on the photo.  The effect of light is merely to provide a scale factor to the resulting image.  In this case, a photograph of a photograph will look identical to the original photograph, if it were captured with brighter or dimmer light.

If the light is not distant, then the intensity of light striking the photograph is generally not uniform.  This effect does not drop off very quickly with the distance to the light.  For example, if a source of light is at a 45$^{\circ}$ angle to the photograph, and the distance to it is eight times as large as half the width of the photo, the near side of the photograph will receive about 40\% stronger illumination than the far side. In the supplementary material we consider the effects of the first order term in the variation of lighting on the photo.  
 
\textbf{Implications:} While we analyze a simple sphere, it is clear that the geometric differences between a face and a sphere will only result in mapping a linear gradient in the image into a more complex variation in lighting on the face, creating potentially more higher order terms.
 To confirm the linear gradient in the images, we train a linear regressor between real images and spoof images. The linear regression input is the real image, and the output is the spoof image with the same identity. After learning this linear regressor, we multiply the spoof image by this gradient and retrain our RGB baseline on the OULU dataset (Protocol 1). We observe that the performance dropped from TPR@FPR=10-2 from 0.53 to 0.51, i.e. error rate for the RGB baseline increases by 3.4\%. This shows that a linear gradient is present in OULU and that the baseline method makes use of it to detect spoofs.

Our method of spoof detection makes use of SFSnet. The SFSnet is trained using a reconstruction loss based on a physical model that uses up to second-order harmonics to model lighting.  Therefore, the SFSnet learns to push into the albedo components of the image that cannot arise from such lighting.  Consequently, when a spoof is taken with a gradient of lighting on the photo that creates third-order harmonics, these will show up as artifacts in the albedo.

\section{Conclusion}
We propose to use albedo as a signal for PAD. We inferred that depth and normal maps are similar for spoof images and non-spoof images, while albedo is visually different. We show that albedo is a better cue for detecting spoofs as compared to normal. We add albedo as an additional signal to seven recent datasets  and achieve state-of-the-art results on all of these datasets. We also show that contrast is being used as cue in existing methods, and removing it results in a drastic drop in performances. Finally, we provide theoretical analysis of how albedo can capture double lighting effects in the spoof images. Finally, we want to add that researchers should note that for the responsible development of this technology, it's important to consider issues of potential unfair bias and consider testing for fairness.
\section{ Acknowledgments}
This work is supported[, in part,] by the US Defense
Advanced Research Projects Agency (DARPA) Semantic
Forensics (SemaFor) Program under [award / grant / contract number] HR001120C0124. Any opinions, findings,
and conclusions or recommendations expressed in this material are those of the author and do not necessarily reflect
the views of the DARPA.

{\small
\bibliographystyle{ieee}
\bibliography{egbib}

\begin{thebibliography}{10}\itemsep=-1pt

\bibitem{Liu_2019_CVPR_Workshops}
L.~Ajian, W.~Jun, E.~Sergio, H.~J.~Escalante, T.~Zichang, Y.~Qi, W.~Kai,
  L.~Chi, G.~Guodong, G.~Isabelle, and S.~Z. Li.
\newblock Multi-modal face anti-spoofing attack detection challenge at
  cvpr2019.
\newblock In {\em Proceedings of the IEEE/CVF Conference on Computer Vision and
  Pattern Recognition (CVPR) Workshops}, June 2019.

\bibitem{albuquerque2020generalizing}
I.~Albuquerque, J.~Monteiro, M.~Darvishi, T.~H. Falk, and I.~Mitliagkas.
\newblock Generalizing to unseen domains via distribution matching, 2020.

\bibitem{Jourabloo_2018_ECCV}
J.~Amin, L.~Yaojie, and L.~Xiaoming.
\newblock Face de-spoofing: Anti-spoofing via noise modeling.
\newblock In {\em ECCV}, September 2018.

\bibitem{Atoum2017FaceAU}
Y.~Atoum, Y.~Liu, A.~Jourabloo, and X.~Liu.
\newblock Face anti-spoofing using patch and depth-based cnns.
\newblock {\em 2017 IEEE International Joint Conference on Biometrics (IJCB)},
  pages 319--328, 2017.

\bibitem{Blanz1999AMM}
V.~Blanz and T.~Vetter.
\newblock A morphable model for the synthesis of 3d faces.
\newblock In {\em SIGGRAPH '99}, 1999.

\bibitem{Boulkenafet2015FaceAB}
Z.~Boulkenafet, J.~Komulainen, and A.~Hadid.
\newblock Face anti-spoofing based on color texture analysis.
\newblock {\em IEEE International Conference on Image Processing}, pages
  2636--2640, 2015.

\bibitem{boulkenafet2015face}
Z.~Boulkenafet, J.~Komulainen, and A.~Hadid.
\newblock face anti-spoofing based on color texture analysis, 2015.

\bibitem{OULU_NPU_2017}
Z.~Boulkenafet, J.~Komulainen, L.~Li, X.~Feng, and A.~Hadid.
\newblock {OULU-NPU}: A mobile face presentation attack database with
  real-world variations.
\newblock May 2017.

\bibitem{Chen2020ASF}
T.~Chen, S.~Kornblith, M.~Norouzi, and G.~E. Hinton.
\newblock A simple framework for contrastive learning of visual
  representations.
\newblock {\em ArXiv}, abs/2002.05709, 2020.

\bibitem{6313548}
I.~Chingovska, A.~Anjos, and S.~Marcel.
\newblock On the effectiveness of local binary patterns in face anti-spoofing.
\newblock In {\em 2012 BIOSIG - Proceedings of the International Conference of
  Biometrics Special Interest Group (BIOSIG)}, pages 1--7, 2012.

\bibitem{Chingovska2012OnTE}
I.~Chingovska, A.~Anjos, and S.~Marcel.
\newblock On the effectiveness of local binary patterns in face anti-spoofing.
\newblock {\em 2012 BIOSIG - Proceedings of the International Conference of
  Biometrics Special Interest Group (BIOSIG)}, pages 1--7, 2012.

\bibitem{feng2018prn}
Y.~Feng, F.~Wu, X.~Shao, Y.~Wang, and X.~Zhou.
\newblock Joint 3d face reconstruction and dense alignment with position map
  regression network.
\newblock In {\em ECCV}, 2018.

\bibitem{frolova2004accuracy}
D.~Frolova, D.~Simakov, and R.~Basri.
\newblock Accuracy of spherical harmonic approximations for images of
  lambertian objects under far and near lighting.
\newblock In {\em ECCV}, pages 574--587. Springer, 2004.

\bibitem{Gan20173DCN}
J.~Gan, S.~Li, Y.~Zhai, and C.~Liu.
\newblock 3d convolutional neural network based on face anti-spoofing.
\newblock {\em 2017 2nd International Conference on Multimedia and Image
  Processing (ICMIP)}, pages 1--5, 2017.

\bibitem{Genova2018UnsupervisedTF}
K.~Genova, F.~Cole, A.~Maschinot, A.~Sarna, D.~Vlasic, and W.~Freeman.
\newblock Unsupervised training for 3d morphable model regression.
\newblock {\em CVPR}, pages 8377--8386, 2018.

\bibitem{George2019DeepPB}
A.~George and S.~Marcel.
\newblock Deep pixel-wise binary supervision for face presentation attack
  detection.
\newblock {\em 2019 International Conference on Biometrics (ICB)}, pages 1--8,
  2019.

\bibitem{He2020MomentumCF}
K.~He, H.~Fan, Y.~Wu, S.~Xie, and R.~B. Girshick.
\newblock Momentum contrast for unsupervised visual representation learning.
\newblock {\em CVPR}, pages 9726--9735, 2020.

\bibitem{Hnaff2019DataEfficientIR}
O.~J. H{\'e}naff, A.~Srinivas, J.~Fauw, A.~Razavi, C.~Doersch, S.~Eslami, and
  A.~Oord.
\newblock Data-efficient image recognition with contrastive predictive coding.
\newblock {\em ArXiv}, abs/1905.09272, 2019.

\bibitem{jia2020singleside}
Y.~Jia, J.~Zhang, S.~Shan, and X.~Chen.
\newblock Single-side domain generalization for face anti-spoofing, 2020.

\bibitem{Stehouwer_2020_CVPR}
S.~Joel, J.~Amin, L.~Yaojie, and L.~Xiaoming.
\newblock Noise modeling, synthesis and classification for generic object
  anti-spoofing.
\newblock In {\em CVPR}, June 2020.

\bibitem{Jourabloo2018FaceDA}
A.~Jourabloo, Y.~Liu, and X.~Liu.
\newblock Face de-spoofing: Anti-spoofing via noise modeling.
\newblock In {\em ECCV}, 2018.

\bibitem{khosla2020supervised}
P.~Khosla, P.~Teterwak, C.~Wang, A.~Sarna, Y.~Tian, P.~Isola, A.~Maschinot,
  L.~Ce, and D.~Krishnan.
\newblock Supervised contrastive learning, 2020.

\bibitem{Komulainen2013ContextBF}
J.~Komulainen, A.~Hadid, and M.~Pietik{\"a}inen.
\newblock Context based face anti-spoofing.
\newblock {\em 2013 IEEE Sixth International Conference on Biometrics: Theory,
  Applications and Systems (BTAS)}, pages 1--8, 2013.

\bibitem{lee2005acquiring}
L.~Kuang-Chih, H.~Jeffrey, and K.~D. J.
\newblock Acquiring linear subspaces for face recognition under variable
  lighting.
\newblock {\em IEEE Transactions on pattern analysis and machine intelligence},
  27(5):684--698, 2005.

\bibitem{Laine2017ProductionlevelFP}
S.~Laine, T.~Karras, T.~Aila, A.~Herva, S.~Saito, R.~Yu, H.~Li, and
  J.~Lehtinen.
\newblock Production-level facial performance capture using deep convolutional
  neural networks.
\newblock {\em Proceedings of the ACM SIGGRAPH / Eurographics Symposium on
  Computer Animation}, 2017.

\bibitem{zhang2003face}
Z.~Lie and S.~Dimitris.
\newblock Face recognition under variable lighting using harmonic image
  exemplars.
\newblock In {\em 2003 IEEE Computer Society Conference on Computer Vision and
  Pattern Recognition, 2003. Proceedings.}, volume~1, pages I--I. IEEE, 2003.

\bibitem{Liu20193DFM}
F.~Liu, L.~Tran, and X.~Liu.
\newblock 3d face modeling from diverse raw scan data.
\newblock {\em ICCV}, pages 9407--9417, 2019.

\bibitem{liu2019deep}
Y.~Liu, J.~Stehouwer, A.~Jourabloo, and X.~Liu.
\newblock Deep tree learning for zero-shot face anti-spoofing, 2019.

\bibitem{Mahajan2020DomainGU}
D.~Mahajan, S.~Tople, and A.~Sharma.
\newblock Domain generalization using causal matching.
\newblock {\em ArXiv}, abs/2006.07500, 2020.

\bibitem{Mancini2020TowardsRU}
M.~Mancini, Z.~Akata, E.~Ricci, and B.~Caputo.
\newblock Towards recognizing unseen categories in unseen domains.
\newblock In {\em ECCV}, 2020.

\bibitem{Pan2007EyeblinkbasedAI}
G.~Pan, L.~Sun, Z.~Wu, and S.~Lao.
\newblock Eyeblink-based anti-spoofing in face recognition from a generic
  webcamera.
\newblock {\em 2007 IEEE 11th International Conference on Computer Vision},
  pages 1--8, 2007.

\bibitem{Patel2016SecureFU}
K.~Patel, H.~Han, and A.~Jain.
\newblock Secure face unlock: Spoof detection on smartphones.
\newblock {\em IEEE Transactions on Information Forensics and Security},
  11:2268--2283, 2016.

\bibitem{Peixoto2011FaceLD}
B.~Peixoto, C.~Michelassi, and A.~Rocha.
\newblock Face liveness detection under bad illumination conditions.
\newblock {\em 2011 18th IEEE International Conference on Image Processing},
  pages 3557--3560, 2011.

\bibitem{Pnvr2020SharinGANCS}
K.~Pnvr, H.~Zhou, and D.~Jacobs.
\newblock Sharingan: Combining synthetic and real data for unsupervised
  geometry estimation.
\newblock {\em 2020 (CVPR)}, pages 13971--13980, 2020.

\bibitem{Qin2020LearningMM}
Y.~Qin, C.~Zhao, X.~Zhu, Z.~Wang, Z.~Yu, T.~Fu, F.~Zhou, J.~Shi, and Z.~Lei.
\newblock Learning meta model for zero- and few-shot face anti-spoofing.
\newblock In {\em AAAI}, 2020.

\bibitem{ramamoorthi2001efficient}
R.~Ravi and H.~Pat.
\newblock An efficient representation for irradiance environment maps.
\newblock In {\em Proceedings of the 28th annual conference on Computer
  graphics and interactive techniques}, pages 497--500, 2001.

\bibitem{basri2003lambertian}
B.~Ronen and D.~Jacobs.
\newblock Lambertian reflectance and linear subspaces.
\newblock {\em IEEE transactions on pattern analysis and machine intelligence},
  25(2):218--233, 2003.

\bibitem{Sengupta2018SfSNetLS}
S.~Sengupta, A.~Kanazawa, C.~D. Castillo, and D.~Jacobs.
\newblock Sfsnet: Learning shape, reflectance and illuminance of faces 'in the
  wild'.
\newblock {\em CVPR}, pages 6296--6305, 2018.

\bibitem{Shu2017NeuralFE}
Z.~Shu, E.~Yumer, S.~Hadap, K.~Sunkavalli, E.~Shechtman, and D.~Samaras.
\newblock Neural face editing with intrinsic image disentangling.
\newblock {\em (CVPR)}, pages 5444--5453, 2017.

\bibitem{Siddiqui2016FaceAW}
T.~A. Siddiqui, S.~Bharadwaj, T.~I. Dhamecha, A.~Agarwal, M.~Vatsa, R.~Singh,
  and N.~Ratha.
\newblock Face anti-spoofing with multifeature videolet aggregation.
\newblock {\em 2016 23rd International Conference on Pattern Recognition
  (ICPR)}, pages 1035--1040, 2016.

\bibitem{Tappen2005RecoveringII}
M.~Tappen, W.~Freeman, and E.~Adelson.
\newblock Recovering intrinsic images from a single image.
\newblock {\em IEEE Transactions on Pattern Analysis and Machine Intelligence},
  27:1459--1472, 2005.

\bibitem{Tewari2017MoFAMD}
A.~Tewari, M.~Zollh{\"o}fer, H.~Kim, P.~Garrido, F.~Bernard, P.~P{\'e}rez, and
  C.~Theobalt.
\newblock Mofa: Model-based deep convolutional face autoencoder for
  unsupervised monocular reconstruction.
\newblock {\em 2017 IEEE International Conference on Computer Vision Workshops
  (ICCVW)}, pages 1274--1283, 2017.

\bibitem{Thies2019Face2FaceRF}
J.~Thies, M.~Zollh{\"o}fer, M.~Stamminger, C.~Theobalt, and M.~Nie{\ss}ner.
\newblock Face2face: real-time face capture and reenactment of rgb videos.
\newblock {\em ArXiv}, abs/2007.14808, 2019.

\bibitem{Tran2017RegressingRA}
A.~Tran, T.~Hassner, I.~Masi, and G.~Medioni.
\newblock Regressing robust and discriminative 3d morphable models with a very
  deep neural network.
\newblock {\em CVPR}, pages 1493--1502, 2017.

\bibitem{Tran2018Nonlinear3F}
L.~Tran and X.~Liu.
\newblock Nonlinear 3d face morphable model.
\newblock {\em 2018 IEEE/CVF Conference on Computer Vision and Pattern
  Recognition}, pages 7346--7355, 2018.

\bibitem{Volpi2018GeneralizingTU}
R.~Volpi, H.~Namkoong, O.~Sener, J.~C. Duchi, V.~Murino, and S.~Savarese.
\newblock Generalizing to unseen domains via adversarial data augmentation.
\newblock In {\em NeurIPS}, 2018.

\bibitem{wagner2011toward}
A.~Wagner, J.~Wright, A.~Ganesh, Z.~Zhou, H.~Mobahi, and Y.~Ma.
\newblock Toward a practical face recognition system: Robust alignment and
  illumination by sparse representation.
\newblock {\em IEEE transactions on pattern analysis and machine intelligence},
  34(2):372--386, 2011.

\bibitem{Wang2018ExploitingTA}
Z.~Wang, C.~Zhao, Y.~Qin, Q.~Zhou, and Z.~Lei.
\newblock Exploiting temporal and depth information for multi-frame face
  anti-spoofing.
\newblock {\em ArXiv}, abs/1811.05118, 2018.

\bibitem{7031384}
D.~Wen, H.~Han, and A.~K. Jain.
\newblock Face spoof detection with image distortion analysis.
\newblock {\em IEEE Transactions on Information Forensics and Security},
  10(4):746--761, 2015.

\bibitem{Wen2015FaceSD}
D.~Wen, H.~Han, and A.~K. Jain.
\newblock Face spoof detection with image distortion analysis.
\newblock {\em IEEE Transactions on Information Forensics and Security},
  10:746--761, 2015.

\bibitem{Chen2020ImprovedBW}
X.Chen, H.Fan, R.~B. Girshick, and K.~He.
\newblock Improved baselines with momentum contrastive learning.
\newblock {\em ArXiv}, abs/2003.04297, 2020.

\bibitem{Yang2019FaceAM}
X.~Yang, W.~Luo, L.~Bao, Y.~Gao, D.~Gong, S.~Zheng, Z.~Li, and W.~Liu.
\newblock Face anti-spoofing: Model matters, so does data.
\newblock {\em (CVPR)}, pages 3502--3511, 2019.

\bibitem{Yu2020MultiModalFA}
Z.~Yu, Y.~Qin, X.~Li, Z.~Wang, C.~Zhao, Z.~Lei, and G.~Zhao.
\newblock Multi-modal face anti-spoofing based on central difference networks.
\newblock {\em CVPR Workshop}, pages 2766--2774, 2020.

\bibitem{Yu2020AutoFasSL}
Z.~Yu, Y.~Qin, X.~Xu, C.~Zhao, Z.~Wang, Z.~Lei, and G.~Zhao.
\newblock Auto-fas: Searching lightweight networks for face anti-spoofing.
\newblock {\em ICASSP 2020 - 2020 IEEE International Conference on Acoustics,
  Speech and Signal Processing (ICASSP)}, pages 996--1000, 2020.

\bibitem{Yu2021NASFASSC}
Z.~Yu, J.~Wan, Y.~Qin, X.~Li, S.~Li, and G.~Zhao.
\newblock Nas-fas: Static-dynamic central difference network search for face
  anti-spoofing.
\newblock {\em IEEE Transactions on Pattern Analysis and Machine Intelligence},
  43:3005--3023, 2021.

\bibitem{CelebA-Spoof}
Z.~Yuanhan, Y.~Zhenfei, L.~Yidong, Y.~Guojun, Y.~Junjie, S.~Jing, and L.~Ziwei.
\newblock Celeba-spoof: Large-scale face anti-spoofing dataset with rich
  annotations.
\newblock In {\em ECCV}, 2020.

\bibitem{qin2019learning}
Q.~Yunxiao, Z.~Chenxu, Z.~Xiangyu, W.~Zezheng, Y.~Zitong, F.~Tianyu, Z.~Feng,
  S.~Jingping, and L.~Zhen.
\newblock Learning meta model for zero-and few-shot face anti-spoofing.
\newblock In {\em AAAI}, 2020.

\bibitem{wang2020deep}
W.~Zezheng, Y.~Zitong, Z.~Chenxu, Z.~Xiangyu, Q.~Yunxiao, Z.~Qiusheng, Z.~Feng,
  and L.~Zhen.
\newblock Deep spatial gradient and temporal depth learning for face
  anti-spoofing.
\newblock In {\em CVPR}, 2020.

\bibitem{Wang_2020_CVPR}
W.~Zezheng, Y.~Zitong, Z.~Chenxu, Z.~Xiangyu, Q.~Yunxiao, Q.~Zhou, Z.~Feng, and
  L.~Zhen.
\newblock Deep spatial gradient and temporal depth learning for face
  anti-spoofing.
\newblock In {\em CVPR}, June 2020.

\bibitem{Zhang2020FaceAV}
K.~Zhang, T.~Yao, J.~Zhang, Y.~Tai, S.~Ding, J.~Li, F.~Huang, H.~Song, and
  L.~Ma.
\newblock Face anti-spoofing via disentangled representation learning.
\newblock {\em ArXiv}, abs/2008.08250, 2020.

\bibitem{zhang2020casiasurf}
S.~Zhang, A.~Liu, J.~Wan, Y.~Liang, G.~Guo, S.~Escalera, J.~Escalante, and
  L.~S. Z.
\newblock Casia-surf: A large-scale multi-modal benchmark for face
  anti-spoofing, 2020.

\bibitem{Zhang2012AFA}
Z.~Zhang, J.~Yan, S.~Liu, Z.~Lei, D.~Yi, and S.~Li.
\newblock A face antispoofing database with diverse attacks.
\newblock {\em 2012 5th IAPR International Conference on Biometrics (ICB)},
  pages 26--31, 2012.

\bibitem{6199754}
Z.~Zhang, J.~Yan, S.~Liu, Z.~Lei, D.~Yi, and S.~Z. Li.
\newblock A face antispoofing database with diverse attacks.
\newblock In {\em 2012 5th IAPR International Conference on Biometrics (ICB)},
  pages 26--31, 2012.

\bibitem{Zhu_2021_CVPR}
X.~Zhu, H.~Wang, H.~Fei, Z.~Lei, and S.~Z. Li.
\newblock Face forgery detection by 3d decomposition.
\newblock In {\em Proceedings of the IEEE/CVF Conference on Computer Vision and
  Pattern Recognition (CVPR)}, pages 2929--2939, June 2021.

\bibitem{yu2020searching}
Y.~Zitong, Z.~Chenxu, W.~Zezheng, Q.~Yunxiao, S.~Zhuo, L.~Xiaobai, Z.~Feng, and
  Z.~Guoying.
\newblock Searching central difference convolutional networks for face
  anti-spoofing.
\newblock In {\em CVPR}, 2020.

\bibitem{yu2020multi}
Y.~Zitong, Q.~Yunxiao, L.~Xiaobai, W.~Zezheng, Z.~Chenxu, L.~Zhen, and
  Z.~Guoying.
\newblock Multi-modal face anti-spoofing based on central difference networks.
\newblock In {\em CVPR Workshop}, 2020.

\bibitem{liu2015faceattributes}
L.~Ziwei, L.~Ping, W.~Xiaogang, and T.~Xiaoou.
\newblock Deep learning face attributes in the wild.
\newblock In {\em (ICCV)}, December 2015.

\end{thebibliography}
}

\end{document}


\title{ Improved Presentation Attack Detection Using Image Decomposition
}
\newcommand\blfootnote[1]{%
  \begingroup
  \renewcommand\thefootnote{}\footnote{#1}%
  \addtocounter{footnote}{-1}%
  \endgroup
}

\author{Shlok Kumar Mishra  \textsuperscript{ \rm 1,\rm 2*}
Kuntal Sengupta  \textsuperscript{\rm 2},
Wen-Sheng Chu  \textsuperscript{\rm 2},\\
Max Horowitz-Gelb  \textsuperscript{\rm 2},
Sofien Bouaziz  \textsuperscript{\rm 2},
David Jacobs\textsuperscript{\rm 1} 
\\
\textsuperscript{\rm 1}University of Maryland, College Park,
    \textsuperscript{\rm 2}Google Research,\\
\texttt{\{shlokm,dwj\}@cs.umd.edu} 
}

\maketitle
\thispagestyle{empty}
\maketitle
\section{Additional Visual Analysis}
In this section we show some additional visual analysis on different datasets.
\blfootnote{* Work was done when Shlok was interning at Google Research.}
\subsection{Visual Analysis on OULU}
\begin{figure}[h!]
  \centering
  \begin{subfigure}[b]{0.4\linewidth}
    \includegraphics[width=\linewidth]{latex/images/real_oulu_1.png}
    \caption{Real Image}
  \end{subfigure}
  \begin{subfigure}[b]{0.4\linewidth}
    \includegraphics[width=\linewidth]{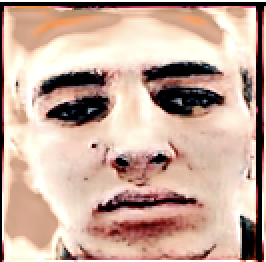}
    \caption{Albedo}
  \end{subfigure}
  \begin{subfigure}[b]{0.4\linewidth}
    \includegraphics[width=\linewidth]{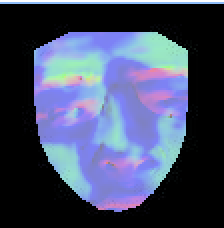}
    \caption{Normal}
  \end{subfigure}
  \begin{subfigure}[b]{0.4\linewidth}
    \includegraphics[width=\linewidth]{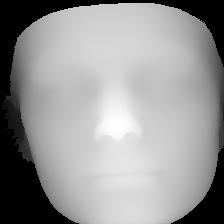}
    \caption{Depth map}
  \end{subfigure}
  \caption{Real Image and the corresponding Albedo and Normal.}
  \label{fig:oulu1}
\end{figure}

\begin{figure}[h!]
  \centering
  \begin{subfigure}[b]{0.4\linewidth}
    \includegraphics[width=\linewidth]{latex/images/spoof_oulu.png}
    \caption{Spoof Image}
  \end{subfigure}
  \begin{subfigure}[b]{0.4\linewidth}
    \includegraphics[width=\linewidth]{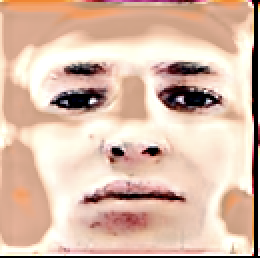}
    \caption{Albedo}
  \end{subfigure}
  \begin{subfigure}[b]{0.4\linewidth}
    \includegraphics[width=\linewidth]{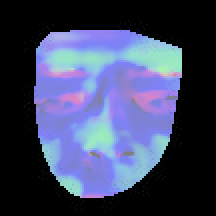}
    \caption{Normal}
  \end{subfigure}
  \begin{subfigure}[b]{0.4\linewidth}
    \includegraphics[width=\linewidth]{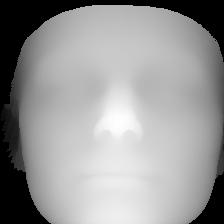}
    \caption{Depth map}
  \end{subfigure}
  \caption{Image and the corresponding Albedo and Normal.}
  \label{fig:oulu2}
\end{figure}
We show some more visual analysis for OULU \cite{OULU_NPU_2017} in Fig \ref{fig:oulu1} and Fig \ref{fig:oulu2}. We can see that difference in albedo in case of real images and spoof images.

\subsection{Visual analysis on CASIA-SURF:}
We have shown some visual analysis for CASIA-SURF \cite{zhang2020casiasurf} as well. We can see that the decomposition not really look great for CASIA-SURF and one of the reasons could be the low quality of images. 

\subsection{Visual analysis on CelebA-Spoof:}
\begin{figure}[t]
  \centering
  \newlength\ww \setlength{\ww}{60pt}
  \begin{tabular}{@{}c@{}c@{}c@{}c@{}}
    Image & Albedo & Normal & Depth  \\
    \includegraphics[width=\ww]{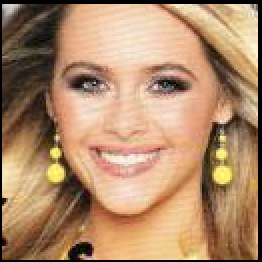} &
    \includegraphics[width=\ww]{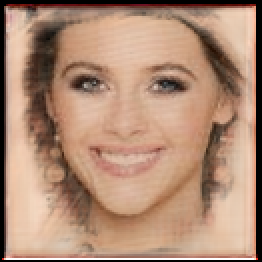} &
    \includegraphics[width=\ww]{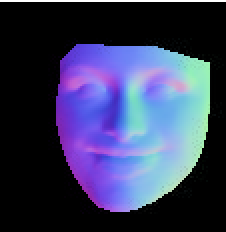} & 
    \includegraphics[width=\ww]{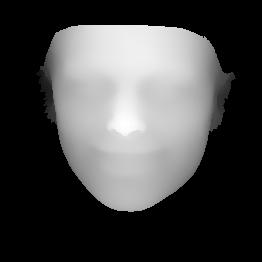} \\
    
    \includegraphics[width=\ww]{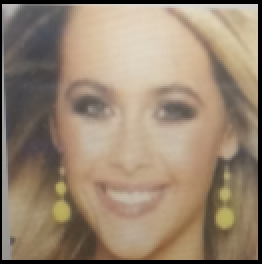} &
    \includegraphics[width=\ww]{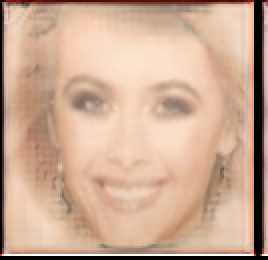} &
    \includegraphics[width=\ww]{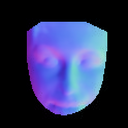} & 
    \includegraphics[width=\ww]{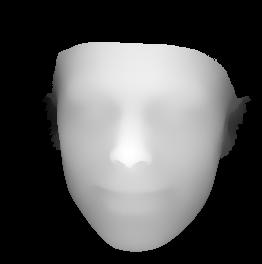} \\
  \end{tabular}
  \caption{
  Live images ({\em top row}) and spoof images ({\em bottom row}) show similar reconstructed depth maps and surface normal.
  Albedo, on the other hand, looks different for spoof images and live images. Albedo helps in capturing the contrast difference resulting from different lighting interactions that happen on the spoof images. 
  Here, we propose to use albedo as a primary cue for detecting spoofs instead of using depth maps.}
  \label{fig:celeba2}
\end{figure}

We have also shown two more visual analysis on CelebA-Spoof Images in Fig \ref{fig:celeba2}.
\begin{figure*}[h!]
  \centering
  \begin{subfigure}[b]{\linewidth}
  \centering
    \includegraphics[width=0.8\linewidth]{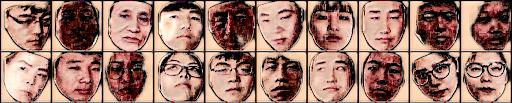}
    \caption{Real Images CASIA-SURF \cite{zhang2020casiasurf}}
  \end{subfigure}
  \begin{subfigure}[b]{\linewidth}
  \centering
    \includegraphics[width=0.8\linewidth]{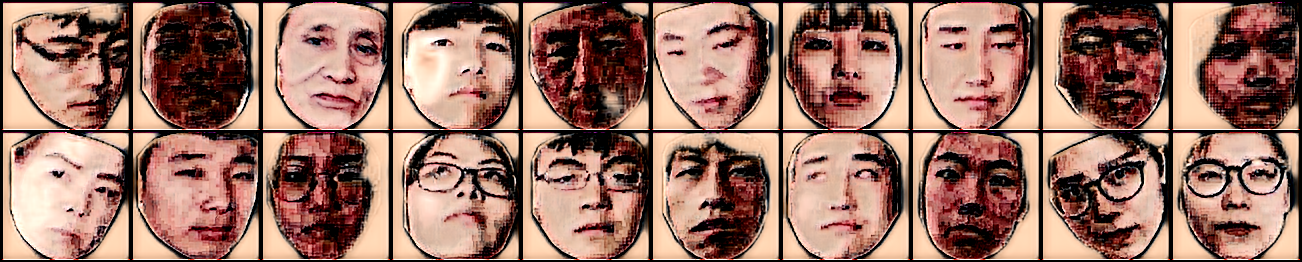}
    \caption{Albedo Real Images CASIA-SURF \cite{zhang2020casiasurf}}
  \end{subfigure}
  \begin{subfigure}[b]{\linewidth}
  \centering
    \includegraphics[width=0.8\linewidth]{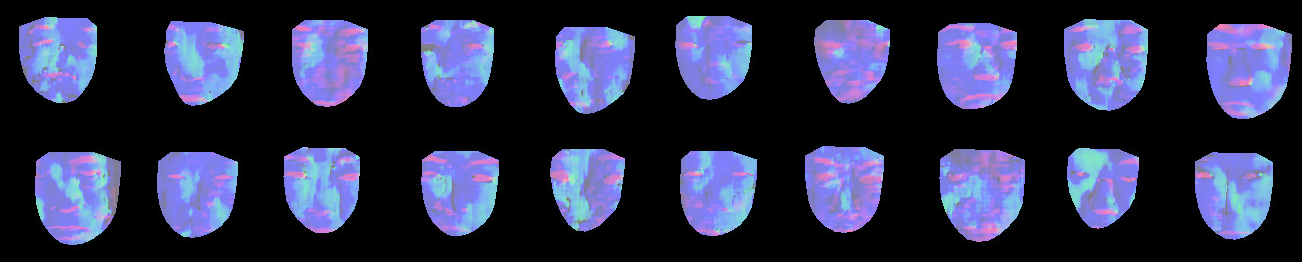}
    \caption{Normal Real Images CASIA-SURF \cite{zhang2020casiasurf}}
  \end{subfigure}
  
  \begin{subfigure}[b]{\linewidth}
  \centering
    \includegraphics[width=0.8\linewidth]{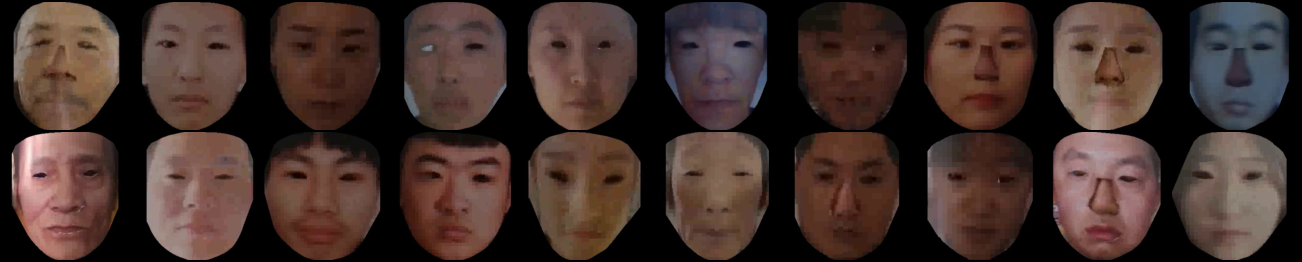}
    \caption{Spoof Images CASIA-SURF \cite{zhang2020casiasurf}}
  \end{subfigure}
  \begin{subfigure}[b]{\linewidth}
  \centering
    \includegraphics[width=0.8\linewidth]{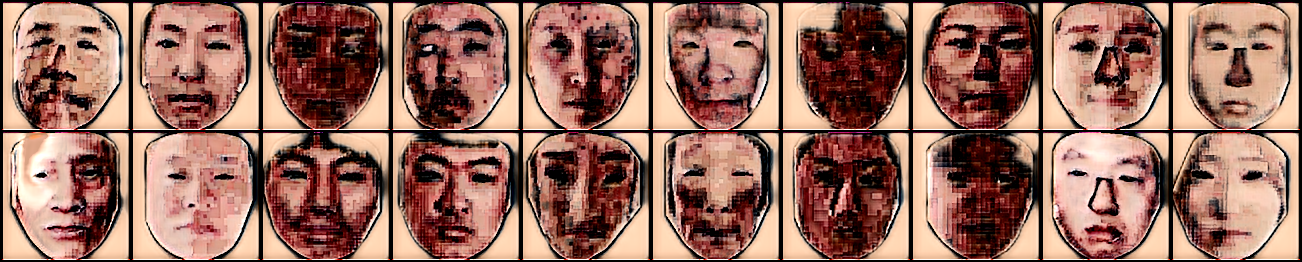}
    \caption{Albedo Spoof Images CASIA-SURF \cite{zhang2020casiasurf}}
  \end{subfigure}
  \begin{subfigure}[b]{\linewidth}
  \centering
    \includegraphics[width=0.8\linewidth]{latex/images/casia_spoof_normal.png}
    \caption{Normal Spoof Images CASIA-SURF \cite{zhang2020casiasurf}}
  \end{subfigure}
  
  \caption{CASIA-SURF Images and the corresponding Albedo and Normal. We can see low quality of images, which makes CASIA-SURF dataset unsuitable for re-training SharinGAN \cite{Pnvr2020SharinGANCS} only on real images to take advantage of domain gap. }
  \label{fig:casia}
\end{figure*}

\section{Additional experiments}
We show additional improved results by adding our method on DCN\cite{Zhang2021StructureDA} and also on the CASIA-SURF HiFiMask dataset. 
\subsection{Comparison with DCN}
 Our proposed method is agnostic to the backbone and can be adapted to any state-of-the-art network. Apart from using CDCN, we have also shown results with DCN by adding albedo component to DCN and have shown state-of-the-art results on OULU dataset in Table \ref{tab:OULU}, on SiW Table \ref{tab:SiW} and cross testing results on CASIA and Replay, Table \ref{tab:CASIA}.
\subsection{Results on CASIA-SURF HiFiMask dataset}
To elaborate on the effect of the albedo maps in challenging attacks such as 3D masks, we show additional improved results on CASIA-SURF HiFiMask dataset. As shown in Table \ref{tab:3D_Masks}, we observe that the effect of lighting on 3D masks becomes even more prominent due to third-order harmonics, which show up as artifacts in the albedo.

\section{Results after contrast normalization:}
\begin{figure}[h!]
  \centering
  \begin{tabular}{@{}cc@{}} 
    \includegraphics[width=0.2\textwidth]{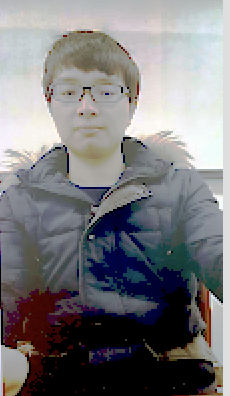} & \includegraphics[width=0.2\textwidth]{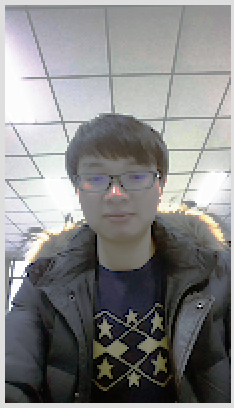} \\
    (a) Spoof Image & (b) Real Image
  \end{tabular}
  \caption{Images after histogram equalization.
  }
  \label{fig:histogram}
\end{figure}

In this section we show visual results after using contrast normalization on OULU images. We can see from Fig \ref{fig:histogram} that images look pretty similar for real images and for spoof images.

We also show the full results on all protocols in Table \ref{Tab:histogram} after doing contrast normalization.

\begin{table}
\centering
\caption{The results of intra testing on three protocols of SiW. } 
\resizebox{0.45\textwidth}{!}{
\begin{tabular}{|c|c|c|c|c|}
\hline
Prot. & Method & APCER(\%) & BPCER(\%) & ACER(\%) \\
\hline
        
        1&\textbf{DCN}&0.0 &0.0 & {0.0} \\
        &\textbf{DCN + SharinGAN(Ours)}&0.0 &0.0 & \textbf{0.0} \\
\hline
       
      2 &\textbf{DCN } &0.00$\pm$0.00 &0.00$\pm$0.0  & \textbf{0.0$\pm$0.0} \\
      &\textbf{DCN + SharinGAN(Ours)} &0.00$\pm$0.00 &0.0$\pm$0.0  & \textbf{0.0$\pm$0.0} \\
\hline
       
       3&\textbf{DCN} &3.80$\pm$4.3 &1.87$\pm$0.10  & 1.70$\pm$0.1 \\
       &\textbf{DCN + SharinGAN(Ours)} &1.13$\pm$0.24 &1.27$\pm$0.11  & \textbf{1.28$\pm$0.13} \\
\hline
\end{tabular}
}
\label{tab:SiW}
\vspace{-0.5em}
\end{table}

\begin{table}
\centering
\caption{The results of intra testing on OULU-NPU. } 
\resizebox{0.45\textwidth}{!}{
\begin{tabular}{|c|c|c|c|c|}
\hline
Prot. & Method & APCER(\%) & BPCER(\%) & ACER(\%) \\
\hline
    1&\textbf{DCN}&1.3 &0.0 & {0.6} \\
    &\textbf{DCN + SharinGAN(Ours)}&0.8 &0.0 & \textbf{0.4} \\
    \hline 
    
    2 &\textbf{DCN } &2.20$\pm$0.00 &2.20$\pm$0.0  & {2.2$\pm$0.0} \\
    &\textbf{DCN + SharinGAN(Ours)} &1.10$\pm$0.00 &1.2$\pm$0.0  & \textbf{1.1$\pm$0.0} \\
    \hline
    
    3&\textbf{DCN} &2.30$\pm$2.7 &1.40$\pm$2.6  & 1.90$\pm$1.6 \\
    &\textbf{DCN + SharinGAN(Ours)} &1.4$\pm$1.3 &1.3$\pm$1.1 & \textbf{1.3$\pm$1.2} \\
    \hline
    
    4&\textbf{DCN} &6.70$\pm$6.8 &0.00$\pm$0.0  & 3.30$\pm$3.4 \\
    &\textbf{DCN + SharinGAN(Ours)} &5.3$\pm$5.5 &0.0$\pm$0.0  & \textbf{2.65$\pm$2.75} \\
\hline
\end{tabular}
}
\label{tab:OULU}
\vspace{-0.5em}
\end{table}

\begin{table}
\centering
\caption{The cross-dataset results on CASIA and Replay. } 
\resizebox{0.45\textwidth}{!}{
\begin{tabular}{|c|c|c|c|c|}
\hline
Method & Train & Test & Train & Test \\
\hline
        &CASIA MFSD & Replay Attack& Replay Attack  & CASIA MFSD  \\
        \hline
        
        DCN & - & 15.3\% & - & 29.4\% \\
        \hline
        
        DCN + SharinGAN(Ours) & -  & \textbf{13.8\%} &  - & \textbf{27.1\%} \\
        \hline
\end{tabular}
}
\label{tab:CASIA}
\vspace{-0.5em}
\end{table}

\begin{table}
\centering
\caption{Results on CASIA-SURF HiFiMask val and test subsets. } 
\resizebox{0.45\textwidth}{!}{
\begin{tabular}{|c|c|c|c|c|}
    \hline
    Method & Evaluation & APCER(\%) & BPCER(\%) & ACER(\%) \\
    \hline
    3DMask\_HRFP \cite{Zhang2021StructureDA}&test &3.78 &2.33 & {3.05} \\
    \hline
    Ours &test& 2.45& 1.27 & {1.86} \\
    \hline
    3DMask\_HRFP \cite{Zhang2021StructureDA}&val &0.85 &1.25 & {1.05} \\
    \hline
    Ours & val &0.54 &0.91 & {0.72} \\
    \hline
\end{tabular}
}
\label{tab:3D_Masks}
\vspace{-0.5em}
\end{table}

\begin{table*}
    \caption{Results after using contrast equalization.}
    \vspace{-1.5ex}
    \centering
    \begin{tabu} to \linewidth{l *{3}{X[c 1.3]}*{2}{X[c 0.8]}}
        \toprule
        {\bf Method} & Protocol & APCER & BPCER & ACER  \\
        \midrule
        CDCN  \cite{yu2020searching} & 1 & 0.4 & 0.0 & 0.2 
        \\
        CDCN contrast removed  \cite{yu2020searching} & 1 & 4 & 3 & 3.5 \\
        
        \midrule
        CDCN  \cite{yu2020searching} & 2 & 1.8 & 0.8 & 1.3 
        \\
        CDCN contrast removed  \cite{yu2020searching} & 2 & 4.9 & 4.5 & 4.7 \\
       
        \midrule
        CDCN  \cite{yu2020searching} & 3 & 1.7 \pm 1.5 & 2.0 \pm 1.2 & 1.8 \pm 0.7 
        \\
        CDCN contrast removed  \cite{yu2020searching} & 3 & 2.64 \pm 0.1  & 13.8 \pm 2.07 & 8.26 \pm 1.02 \\
        
        \midrule
        CDCN  \cite{yu2020searching} & 4 & 4.2 \pm 3.4 & 5.8 \pm 4.9 & 5.0 \pm 2.9 
        \\
        CDCN contrast removed  \cite{yu2020searching}  & 4 & 7.5 \pm 2.5 & 15 \pm 4.9 & 11.25 \pm 1.25  \\
        \bottomrule
    \end{tabu}
    \caption{We show the impact of removing contrast as cue from spoof detection systems. We can see that after removing contrast we see performance drop in spoof detection on CDCN\cite{yu2020searching}. Hence to capture this contrast cue and other lighting cues we propose to use Albedo to detect spoof images.}
    \label{Tab:histogram}
    
\end{table*}

 \subsection{Impact of \#parameters on the performance:}
To analyze whether the performance gain is due to adding additional backbone (thus increasing parameters in network) or due to using albedo information; we drop the additional albedo component in the CDCN architecture. Now in addition to RGB images we add either the albedo or the depth map in CDCN architecture. Note that for this experiment we don't hardcode the spoof depth map to be zero and use the output or PRNet for both real and spoof images as the depth map.
We find that adding the albedo as additional input helps more than adding depth as additional input as shows in Table \ref{Tab:parameters}. Hence our performance improvement is not just because we have increased the numbers of parameter in the network, but because we use albedo component which helps in detecting better spoof images.
\begin{table}
    \vspace{-0.5ex}
     \caption{We show the performance by using Albedo and depth maps in CDCN architecture. For a fair comparison we are fixing the #parameters in the CDCN architecture. We either use depth as or albedo as additional signal to CDCN architecture. We can see that albedo performs on-par with depth maps on CDCN architecture.}
    \centering
   
    \begin{tabu} to \linewidth{l *{2}{X[c 1.3]}*{2}{X[c 0.8]}}
        \toprule
        {\bf Method} & Protocol & APCER & BPCER & ACER  \\
        \midrule
        Using Depth & 1 & 1.2 & 11.4 & 6.1 
        \\
        Using Albedo & 1 & \textbf{1.1} & \textbf{8.4} & \textbf{4.7}\\
        
        \bottomrule
    \end{tabu}
   
    \label{Tab:parameters}
    
\end{table}
 

 \section{Supervised Contrastive learning advantages:}
 One of the additional added advantages of using supervised contrastive learning is its further extension by adding a face matcher in the loop.  Face recognition systems in practice first try to recognize and match the face and then use the PAD system to detect whether the face is spoof or not. However PAD systems in prior research work \cite{yu2020searching,Yu2020MultiModalFA,jia2020singleside} don't use this to their advantage, they instead directly try to predict spoofs/real images in isolation. To achieve this, we can add a face matcher in the loop. This enables us to pair every spoof image to a real image that matches it with a score above a certain threshold, and are potentially from the same identity. For every positive sample we can ensure that we have a negative sample of the same identity in the same minibatch.

\begin{table*}
    \caption{In this experiment we replace our resnet34 backbone by resnet152 backbone. We can see that even after increasing the number of parameters by close to 2.75 times, final accuracy is still 38.35\% less than resnet34 backbone. Hence this verifies our claim that doing image decomposition explicitly helps the network learn better features. }
    \vspace{-1.5ex}
    \label{tab:CASIA}
    \centering
    \begin{tabu} to \linewidth {lccc} 
        \toprule
        {\bf Method} & TPR@FPR=10-2 & TPR@FPR=10-3 \\ 
        \midrule
        Res34 & 0.38 & 0.08\\ 
        
        Res152 & 0.45 & 0.09 \\
        
        IDCL(Ours) & 0.73 & 0.33 \\
        
        \bottomrule
    \end{tabu}
    
\end{table*}

    
\begin{figure*}[t]
    \centering
    \vspace{-2.5pt}
    \includegraphics[width=\linewidth]{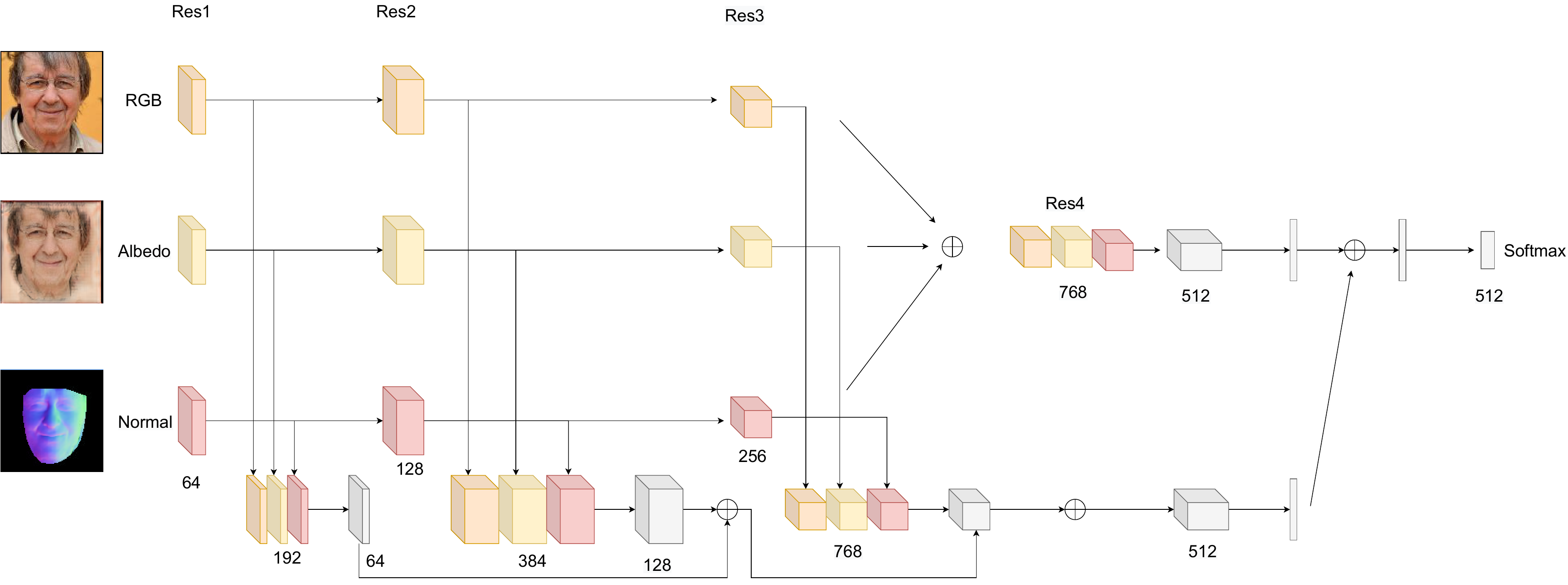}
    \caption{
    {\bf Detailed Architecture diagram (IDCL):} 
    We use image decomposition SharinGAN \cite{Pnvr2020SharinGANCS} to obtain three networks of RGB, normal and albedo images. 
    We use additional aggregation layers to fuse the information in the final layers.}
\end{figure*}

\section{Experiments with deeper networks:} Another hypothesis we want to test is whether the improvement we have seen using image decomposition, could be achieved by only RGB baseline if we train on deeper and bigger networks. To illustate that we replace our resnet34 backbone with resnet152 and results can be seen in Table \ref{tab:CASIA}. We can see that even after increasing the number of parameters by close to 2.75 times, final accuracy is still 38.35\% less than original network. Hence this verifies our claim that doing image decomposition explicitly helps the network learn better features. 

{\small
\bibliographystyle{ieee}
\bibliography{egbib}
}